\documentclass[conference]{IEEEtran}
\IEEEoverridecommandlockouts
\usepackage{cite}
\usepackage{amsmath,amssymb,amsfonts}
\usepackage{algorithmic}
\usepackage{graphicx}
\usepackage{caption}
\usepackage{subcaption}
\usepackage{textcomp}
\usepackage{hyperref}
\usepackage{xcolor}
\def\BibTeX{{\rm B\kern-.05em{\sc i\kern-.025em b}\kern-.08em
    T\kern-.1667em\lower.7ex\hbox{E}\kern-.125emX}}

\begin{document}

\title{Effects of Spectral Normalization in Multi-Agent Reinforcement Learning

\thanks{Code Link: \href{https://github.com/kinalmehta/epymarl\_spectral/}{https://github.com/kinalmehta/epymarl\_spectral/}}
}

\author{\IEEEauthorblockN{Kinal Mehta}
\IEEEauthorblockA{\textit{CSTAR} \\
\textit{IIIT Hyderabad}\\
Hyderabad, India \\
kinal.mehta@research.iiit.ac.in}
\and
\IEEEauthorblockN{Anuj Mahajan}
\IEEEauthorblockA{\textit{Department of Computer Science, } \\
\textit{University of Oxford}\\
Oxford, UK \\
anuj.mahajan@cs.ox.ac.uk}
\and 
\IEEEauthorblockN{Pawan Kumar}
\IEEEauthorblockA{\textit{CSTAR} \\
\textit{IIIT Hyderabad}\\
Hyderabad, India \\
pawan.kumar@iiit.ac.in}
}

\maketitle

\begin{abstract}
A reliable critic is central to on-policy actor-critic learning. But it becomes challenging to learn a reliable critic in a multi-agent sparse reward scenario due to two factors: 1) The joint action space grows exponentially with the number of agents 2) This, combined with the reward sparseness and environment noise, leads to large sample requirements for accurate learning. We show that regularising the critic with spectral normalization (SN) enables it to learn more robustly, even in multi-agent on-policy sparse reward scenarios. Our experiments show that the regularised critic is quickly able to learn from the sparse rewarding experience in the complex SMAC and RWARE domains. These findings highlight the importance of regularisation in the critic for stable learning.
\end{abstract}


\section{Introduction}

Multi-agent reinforcement learning (MARL) framework \cite{mehta2023marljax} can be used to formulate many real-world tasks in autonomous driving, robotics, etc. Having multiple agents introduces several new challenges \cite{mahajan2022generalization}, which include exponential growth in the joint action space, non-stationarity in the environment due to co-evolving agents, exploration in the joint action space \cite{mahajan2019maven,gupta2020uneven}, credit assignment and gradient variance. Non-stationarity arising from multi-agents is usually dealt with using a centralised training approach. But dealing with all the challenges of exponential growth of the joint action space remains an open problem. 
All these challenges, when combined with even a little sparsity in rewards, make learning very difficult in MARL.
The most successful approach to decentralised cooperative MARL has been centralised training decentralised execution (CTDE) \cite{lowe2017multi}.
Many value-based \cite{mahajan2019maven, qmix, son2019qtran, wang2020rode} and policy-based \cite{yu2021surprising, pmlr-v139-mahajan21a} methods have been developed under the umbrella of CTDE. The method Multi-agent PPO (MAPPO) \cite{yuMAPPOSurprisingEffectivenessPPO2022a} is a widely used on-policy MARL algorithm which is able to match the performance of off-policy value-based methods that have been shown to perform better on various environments. 

Actor-critic algorithms have been successfully used in many single-agent \cite{schulmanPPOProximalPolicyOptimization2017} and multi-agent \cite{yuMAPPOSurprisingEffectivenessPPO2022a, pmlr-v139-mahajan21a} reinforcement learning tasks. Critic is a central part of the actor-critic framework, which evaluates the action produced by the actor. The effectiveness of the actor depends on the effectiveness of the critic; increasing the stability of the critic directly correlates to increasing the actor's effectiveness \cite{sn-large-drl}.
In this work, we focus on the sparse rewards scenarios, which are often seen in real-world settings.
We show that the capability of MAPPO is hampered significantly in sparse reward scenarios. 
To address this, we propose a critic regularisation technique called Spectral Normalization that leads to better critic convergence which in turn leads to better policy convergence. The generative adversarial networks \cite{goodfellow2014} are known to be notoriously difficult to train, and numerous methods have been proposed \cite{florian2019,lars2017conopt,kumar2023}; it involves solving a delicate minmax optimization problem. Spectral normalization has been used to stabilize the discriminator in Generative Adversarial Networks (GANs) \cite{sn-gan}. 
We hypothesize that by introducing sparsity in the reward, critic learning gets affected and propose to regularise critic with spectral normalization to aid the critic to learn. Applying spectral normalization constraints the Lipschitz constant of the layers. 

We empirically study the effects of reward sparsity on critic learning in MAPPO on two different cooperative multi-agent benchmarks: StarCraft multi-agent challenge (SMAC) \cite{smac} and multi-robot warehouse (RWARE) \cite{epymarl}. We start by comparing the performance of MAPPO with critic regularised MAPPO. We then analyze the critic learning by comparing the logarithm of the gradient norms of the critic of the two variants and show how applying spectral normalization on the critic helps stabilize its gradients. Our results help us understand the importance of critic learning in multi-agent scenarios under sparse rewards.

Our contributions can be summarised as follows:
\begin{itemize}
\item We introduce a sparse reward configuration for SMAC and show that it is difficult to learn when compared to standard reward configuration.
\item We propose to regularise the critic with spectral normalization and show that it helps learn better policies under sparse rewards.
\item We analyse the effects of applying spectral normalization and show that it helps 1) stabilise the critic gradients and 2) has an optimization effect of scaling the gradients of the entire critic by the product of the largest spectral value of the weight matrices.
\end{itemize}

We start by introducing the multi-agent RL, PPO, MA-PPO and Spectral Normalization in Section \ref{section: background} followed by discussion on optimization effects of Spectral Normalization in Section \ref{section:optim_effect}. After that we describe our experimental setup in Section \ref{section: experiment} and present our results and discussion in Section \ref{section:results}.

\begin{figure*}[th]
    \centering
    \begin{tabular}{c c}
        \includegraphics[height=0.22\textheight]{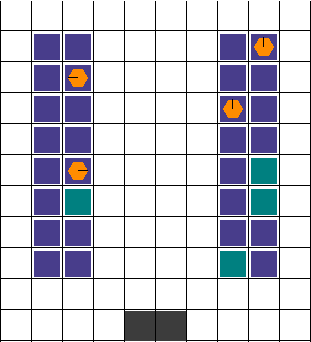} &
        \includegraphics[height=0.22\textheight]{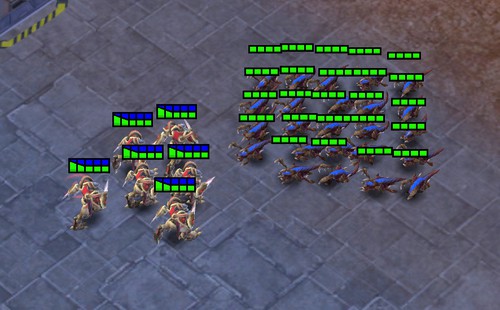} \\
    \end{tabular}
    \caption{Illustration of an RWARE environment and a SMAC map}
    \label{fig:rwareEnv}
\end{figure*}

\section{Related Work}
There has been considerable development in cooperative multi-agent reinforcement learning in recent years\cite{ qmix, sunehagVDNValueDecompositionNetworksCooperative2017, mahajan2019maven, yuMAPPOSurprisingEffectivenessPPO2022a, pmlr-v139-mahajan21a}. Value-based as well as policy-based CTDE-MARL algorithms are effective in cooperative tasks. But these works do not focus on the sparse reward scenarios and mainly address learning decentralized agents with factored value functions or policies.

Spectral Normalization has been used in GANs\cite{sn-gan}, as a regularizer which leads to better sample efficiency \cite{goukRegularisationNeuralNetworks2020} or to improve robustness of uncertainty estimates \cite{liuSimplePrincipledUncertainty2020}. In the context of RL, SN has been used in model-based RL in uncertainty estimation \cite{yuMOPOModelbasedOffline2020} to enable deeper networks \cite{sn-large-drl} and to also show that SN regularised networks can compete with algorithmic innovations \cite{sn-rl-opt}. Tesseract \cite{pmlr-v139-mahajan21a} uses tensor decompositions to learn robust estimates for the underlying MDP dynamics and action-value function with provable sample efficiency in multi agent setting.

To the best of our knowledge, we are the first to apply SN in the context of multi-agent RL. Our work differs from the previous works in the sense that we show that SN can be used to make critic more robust to the noise induced by sparse rewards under multi-agent scenarios. Previous works have shown that adding SN to the value function estimator helps stabilize its learning by stabilizing its gradients \cite{sn-large-drl} as well as act as an update-step scheduler \cite{sn-rl-opt}. In our work, we observe that applying SN in multi-agent scenarios leads to both of these benefits.

\section{Background}
\label{section: background}
\subsection{Cooperative MARL}
We consider a fully cooperative multi-agent reinforcement learning task and model it as a decentralized partially observable MDP (Dec-POMDP). Dec-POMDP can be defined by tuple $\{S, U, P, r, Z, O, n, \gamma\},$ where $S$ is the state space of the environment, and $z^i \in Z$ is the local observation of each agent sampled according to the observation function $O(s,i): S \times \mathcal{A} \rightarrow Z$.  
The action-observation history for an agent $i$ is $\tau^i \in T \equiv (Z \times U)^*$, on which the policy $\pi^i(u^i|\tau^i) : T \times U \rightarrow [0,1]$ of each agent is conditioned.  
At each time step $t$, every agent $i \in \mathcal{A} \equiv \{1, \, \dots, \, n\}$ chooses an action $u^i \in U$ with a decentralised policy $\pi^i(\cdot|\tau^i)$ using only its local action-observation history $\tau^i$.  
The agents jointly optimize the discounted accumulated reward 
$$J = \mathbb{E}_{s_t, \mathbf{u}_t} \left[ \sum_t \gamma^t r(s, \mathbf{u}) \right],$$ 
where the joint action space $\mathbf{u} \in \mathbf{U} \equiv U^n$ can be denoted as a tuple $\mathbf{u} = (u^1, \dots, u^n)$. When $n=1$ the problem becomes a POMDP and is significantly easier to solve.  
Here $P(s'|s,\mathbf{u}): S \times \mathbf{U} \times S \rightarrow [0,1]$ is the state transition function,
$r(s, \mathbf{u}) : S \times \mathbf{U} \rightarrow \mathbb{R}$ is the reward function shared by all agents and $\gamma \in [0,1)$ is the discount factor. 
The state-value function conditioned on joint policy $\boldsymbol{\pi}$ is defined as 
$$V^{\boldsymbol{\pi}}(s_t) = \mathbb{E}_{\mathbf{u} \sim \boldsymbol{\pi}} \left[ \sum_{k=0}^{\infty} \gamma^k r_{t+k} \, | \, \mathbf{s} \right].$$
A collaborative team aims to learn an optimal joint policy $\boldsymbol{\pi} = \Pi_{i=1}^{n}\pi^i$ which maximizes the accumulated reward $J$.

\subsection{PPO and MA-PPO}
PPO \cite{schulmanPPOProximalPolicyOptimization2017} is a single-agent actor-critic algorithm which optimizes the clipped objective with a $KL$ penalty.
The objective for policy optimization under PPO is
\begin{equation}
    \mathbb{E}_t[\min(\rho_t(\theta)A_t, \text{clip}(\rho_t(\theta),1-\epsilon,1+\epsilon)A_t) - \beta \cdot KL_p],
\end{equation}
where $A_t$ is the advantage for that given state $s_t$ and action $u_t$, $\theta$ is the policy network weights, $\theta_{old}$ is the policy network weights using which the action was selected, %
$$\rho_t(\theta) = \dfrac{\pi_\theta(u_t|s_t)}{\pi_{\theta_{old}}(u_t|s_t)}$$ 
is the probability ratio of the selected action and $$KL_p = KL[\pi_{\theta_{old}}(\cdot|s_t), \pi_\theta(\cdot|s_t)]$$ is the $KL$ divergence between the old and the new policy distributions.
The advantage is calculated as follows.  
\begin{equation}
    A_t(s_t,u_t) = r_t + \gamma \cdot V^\pi_\phi(s_{t+1}) - V^\pi_\phi(s_t),
\end{equation}
where $V_\pi$ is the value function or critic.  
The critic is trained to minimise the following objective.
\begin{equation}
    \min_{\phi} \, [G_t - V^\pi_\phi(s_t)]^2.
\end{equation}
Here, the training of the policy is driven by the value prediction accuracy from the critic.

MAPPO \cite{yuMAPPOSurprisingEffectivenessPPO2022a} is a multi-agent extension of PPO where the critic is centralised and has access to privileged information during the training. The centralised critic learns the joint value function of the cooperative environment.

In a multi-agent scenario, the challenges of training critic are even more severe due to an exponential blowup of the joint action space \cite{pmlr-v139-mahajan21a} and potential non-stationarity of the environment. When using a central critic with a CTDE framework, a single critic with the same set of parameters is responsible for learning the value prediction for all the agents. This might lead to conflicting goals for the critic leading to unstable critic updates. This problem amplifies with the increase in number of agents.

As we mentioned earlier, the learning of policy depends on how accurate the critic's predictions of value estimates are. When we introduce one more challenge of reward sparsity, the noise in critic learning is further increased due to bootstrapping from an inaccurate critic. The problem of an unstable critic is even more prominent when the rewards become sparse. In complex scenarios like SMAC, this could also cause the actor-critic based agents to not even find the optimal policy, which is reflected in our results in Fig. \ref{fig:smac_battles_won}. In a sparse reward setting, there is very little signal from the environment to improve the value prediction and hence it becomes difficult for the critic to learn the actual state-value function. Techniques like HER \cite{andrychowiczHindsightExperienceReplay} have been developed for off-policy learning.

\subsection{Spectral Normalization in Reinforcement Learning}
Spectral normalization (SN) has been used to stabilise the discriminator learning in GANs\cite{miyatoSpectralNormalizationGenerative2022}.
A function is $k-$Lipschitz continuous in $l_2$-norm if   
\begin{equation}
    \lVert f(x_1) - f(x_2) \rVert_2 \le k \lVert x_1 - x_2 \rVert_2.
\end{equation}
Considering a feed-forward layer, the Lipschitz constant of the layer is defined as the largest singular value of the weight matrix of that layer.  
Spectral normalization normalizes the weight matrix by its largest spectral value, constraining that layer to be $1-$Lipschitz smooth.
\begin{equation}
    \hat W = \frac{W}{\lVert W \rVert} = \frac{W}{\sigma_{\max}(W)}.
\end{equation}
We can also control the smoothness of the function to be $k$ Lipschitz smooth by adding an extra parameter $k$ which can be tuned.
\begin{equation}
    \hat W = \frac{W}{\operatorname{max}(\sigma_{\max}(W), k)}.
\end{equation}

The Lipschitz constant of a composite of two functions $f_1$ and $f_2$ with Lipschitz constant $k_1$ and $k_2$ will be bounded by $k_1 \cdot k_2$. Similarly the Lipschitz constant of a neural network can be bounded by the product of Lipschitz constant of each layer. For more details on Lipschitz constant of various layers and activation functions, we refer to \cite{goukRegularisationNeuralNetworks2020}.

We can draw parallels between GANs and actor-critic RL algorithms. 
Just as the performance of the generator is driven by the accuracy of the discriminator, in actor-critic, the performance of actor or policy is driven by the accuracy of the critic.
We use spectral normalization in the critic to stabilize its gradients and learning. Using SN makes the critic updates more stable and hence aids learning of the policy. In case of sparse rewards scenario, the noise from the bootstrapped updated usually interferes with the actual reward signal. SN helps mitigate this issue by constraining the layers to be Lipschitz continuous and hence bounding the representation space.

As done in \cite{sn-gan}, the power iteration method is used to compute the largest spectral value for applying spectral normalization. The additional computation cost is relatively small compared to the full computational cost of the critic network. The time comparisons when applying spectral normalization on GANs is shown in \cite{sn-gan}.

\begin{figure*}[th]
    \centering
    \begin{tabular}{c c c}
        \includegraphics[width=0.32\linewidth]{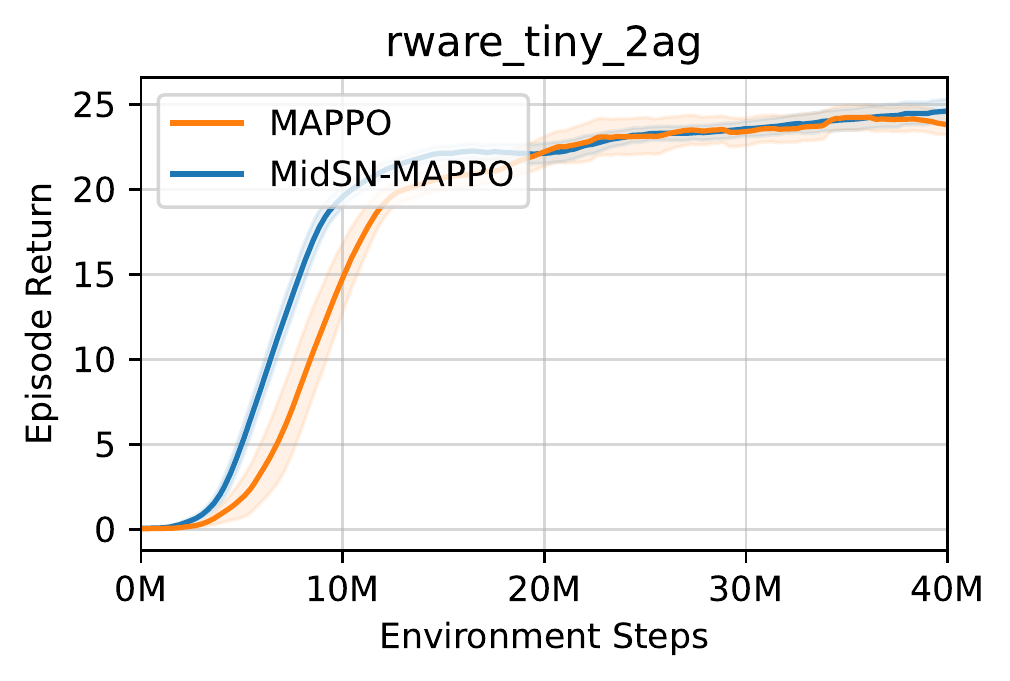} &
        \includegraphics[width=0.32\linewidth]{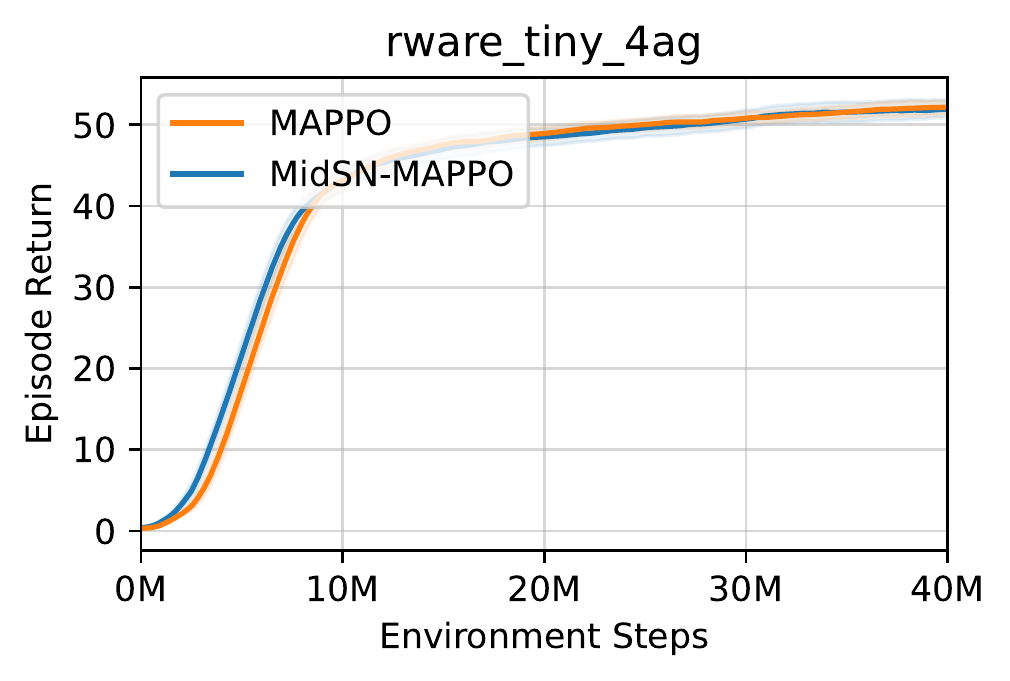} &
        \includegraphics[width=0.32\linewidth]{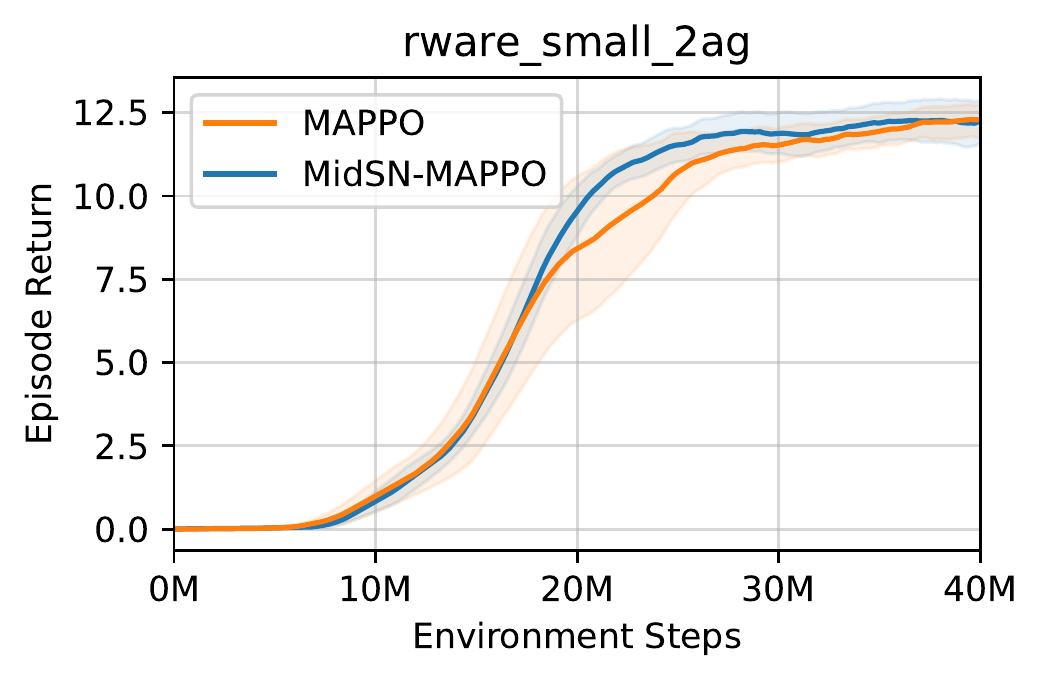}
    \end{tabular}
    \caption{Learning curves on RWARE comparing MAPPO and MidSN-MAPPO. As RWARE is a relatively simple environment where explicit coordination is not necessary, the final performance of both variants is almost the same. However, we can observe that MidSN-MAPPO converges a bit faster.}
    \label{fig:rware}
\end{figure*}

\section{Optimization effects of Spectral Normalization}
\label{section:optim_effect}
Let us analyze the activation calculation of a feed-forward layer with and without spectral normalization. The equation for a layer $i$ without spectral normalization can be written as:
\begin{align}
    z_i &= W_i a_{i-1} + b_i \\
    a_i &= \operatorname{ReLU}(z_i),
\end{align}
where $a_0 \triangleq x$ is the input to the network.

Now let us look at the equations when we apply spectral normalization to a feed-forward layer.
\begin{align}
    \hat{z_i} &= \hat{W_i} a_{i-1} + b_i \\
    \hat{a_i} &= \operatorname{ReLU}(\hat{z_i}),
\end{align}
Here $\hat{W_i} = k_i^{-1} W_i$ is the weight matrix after applying spectral normalization. Here $k_i$ is the largest singular value of the weight matrix.
Comparing the above equations, we can observe that only the weight matrix is scaled using the largest singular value, whereas the bias is unchanged. Due to this, the sign of the pre-activations $z$ is not preserved ($[z_i > 0] \neq [\hat{z_i} > 0]$). Hence we cannot write a direct relation between $\frac{\partial \mathcal{L}}{\partial W_i}$ and $\frac{\partial \hat{\mathcal{L}}}{\partial W_i}$

For simplicity of analysis let us consider the network without bias. So the equation for a specific layer $i$ can be written as follows:
\begin{align}
    z_i &= W_i a_{i-1} \\
    a_i &= \operatorname{ReLU}(z_i),
\end{align}
where $a_0 \triangleq x$ is the input to the network.

Let a subset of layer $\mathcal{S} \subseteq\{1,2, \ldots, L\}$ are spectral normalized and are individually 1-Lipschitz continuous. The weight matrix of the regularised layers can be defined as $\forall i \in \mathcal{S} : \hat{W_i} = \langle k_i^{-1}\rangle W_i$ where $k_i = \sigma_{\max}(W_i)$ is the largest singular value of that weight matrix. Here $\langle \cdot \rangle$ is the gradient stop operator and hence back-propagation is not applied through the singular value calculations.

Now let us update the equations for the above described feed-forward network when applying spectral normalization to it.
\begin{equation} \label{eq:weight_normalized_layer_op}
    \hat{z}_i = k_{i}^{-1} W_i \hat{a}_{i-1}
\end{equation}
\begin{equation}
    \hat{a}_i = \operatorname{ReLU}(\hat{z}_i),
\end{equation}
where $k_{i: j}^{-1} \triangleq \prod_{i \leq l \leq j \wedge l \in \mathcal{S}} k_l^{-1}$.
We can write eq. \ref{eq:weight_normalized_layer_op} in terms of non-regularised activation as follows
\begin{equation}
    \hat{z}_i = k_{1:i}^{-1} W_i {a}_{i-1}.
\end{equation}
The above equation is valid as spectral normalization is scaling operation and hence the sign of the activation will be preserved ($[a_i > 0] = [\hat{a}_i > 0]$).

The loss is calculated on the final layer of the network and hence can be written as $\mathcal{L} \triangleq \operatorname{loss}(z_L)$. The loss calculation for the regularised network will be updated to $\hat{\mathcal{L}} \triangleq \operatorname{loss}(\hat{z}_L) = \operatorname{loss}(k_{1:L}^{-1} z_L)$.
\begin{align}
    &\operatorname{MLP} & &\operatorname{SN-MLP} \\
    \mathcal{L} &\triangleq \operatorname{loss}(z_L) & \hat{\mathcal{L}} &\triangleq \operatorname{loss}(\hat{z}_L) \\
    \frac{\partial \mathcal{L}}{\partial W_i} &= J_i {\delta}_L a_{i-1}^T & \frac{\partial \hat{\mathcal{L}}}{\partial W_i} &= k^{-1} J_i \hat{\delta}_L \hat{a}_{i-1}^T,
\end{align}
where $k^{-1} = \Pi_{i \in \mathcal{S}}k_i^{-1}$, $\delta_{L} \triangleq \frac{\partial \mathcal{L}}{\partial z_L}$ is the Jacobian w.r.t. the network's output and similarly $\hat{\delta}_{L} \triangleq \frac{\partial \hat{\mathcal{L}}}{\partial \hat{z}_L}$ is the Jacobian with respect to the regularised network's output and 
$$J_i \triangleq \Pi_{j=i}^{L-1}[\operatorname{diag}([z_j]>0)W_{j+1}^T].$$

Based on the above equations, it is evident that applying spectral normalization leads to gradient scaling by $k^{-1}$. This shows that the optimization step of the regularised network is scheduled based on the product of largest spectral values of the normalized layers.
For detailed analysis of how spectral normalization effects various layers, activation and the bias terms, we refer to \cite{sn-rl-opt, goukRegularisationNeuralNetworks2020}. We also note here that spectral normalization is a form of preconditioning \cite{calrson2015,katyan2020,Das2021,das2020,kumar2013multi,kumar2014aggregation,kumar2016relaxed,li2015preconditioned,qiao2019}.

Under sparse rewards, the learning of the critic is unstable as it uses bootstrapped targets from an untrained critic. This could lead to unpredictable updates in the weight matrices. When regularizing the critic with SN, the gradient scaling with $k^{-1}$ restricts the model weights from diverging due to incorrect target estimates. While once the critic is trained a bit, it leads to more accurate and consistent bootstrapped targets.

\begin{figure*}[th]
    \centering
    \begin{tabular}{c c c}
        \includegraphics[width=0.32\linewidth]{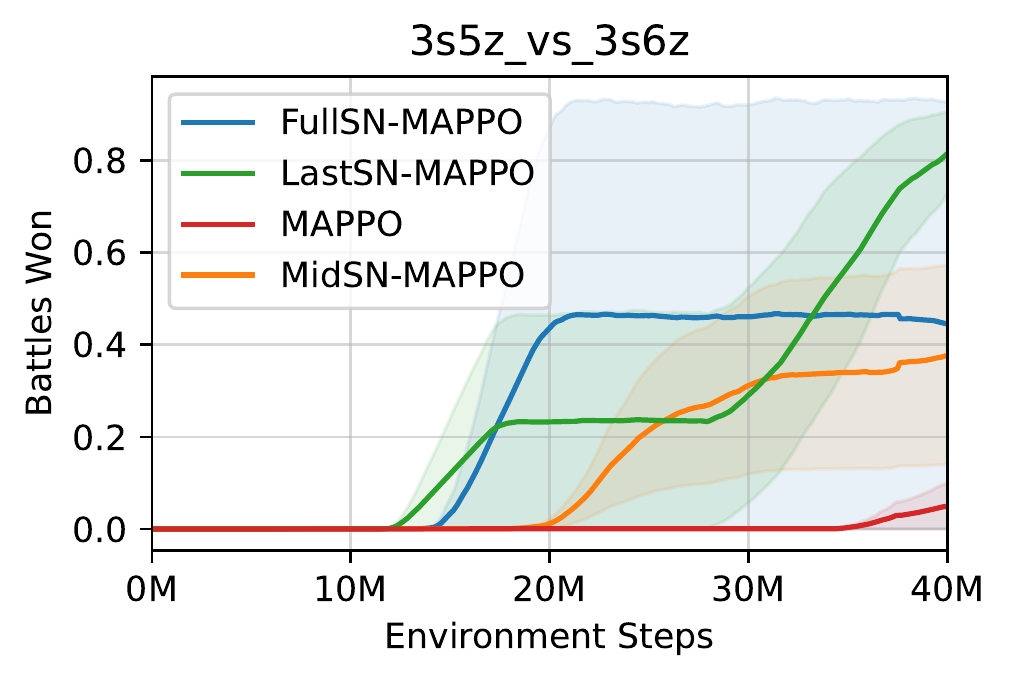} &
        \includegraphics[width=0.32\linewidth]{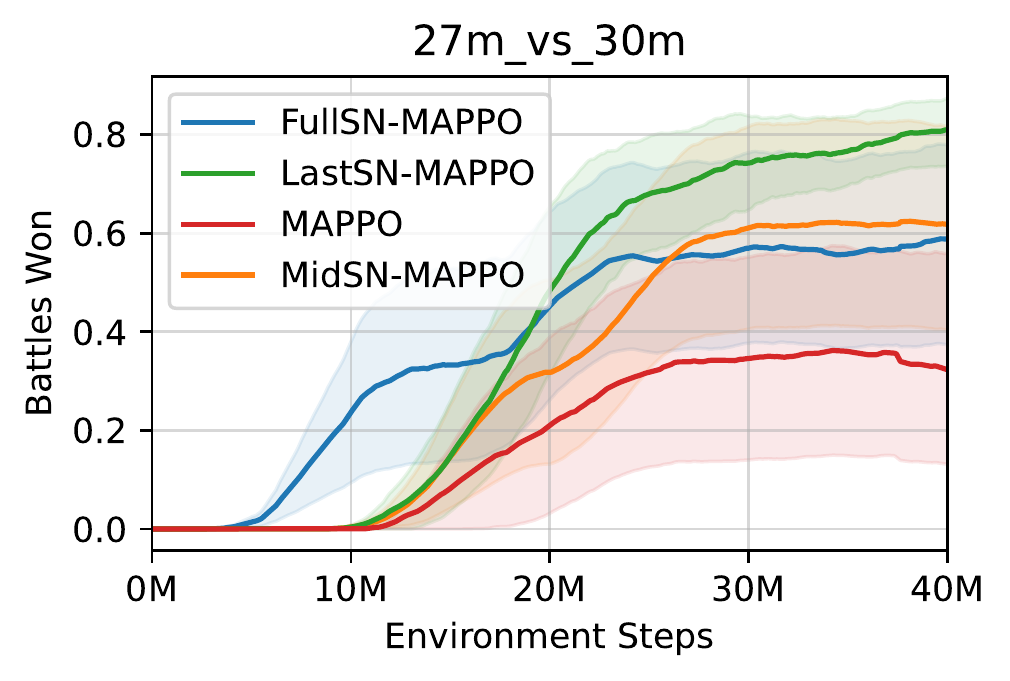} &
        \includegraphics[width=0.32\linewidth]{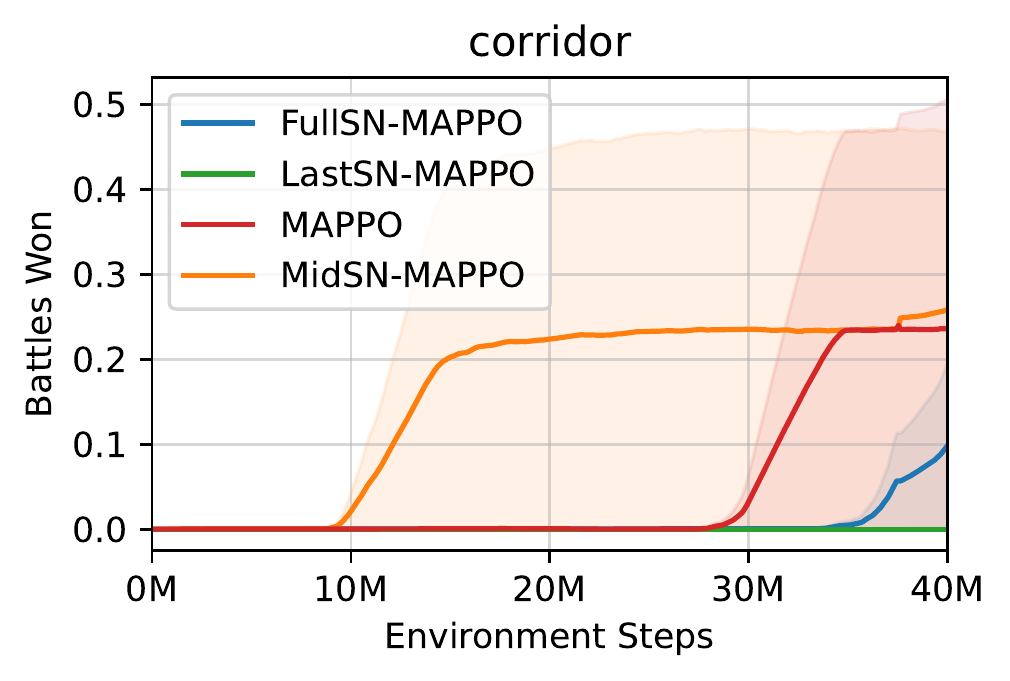}
    \end{tabular}
    \caption{Average battles won on various SMAC maps averaged across several seeds. SMAC is a very challenging benchmark where each map requires a specific skill to be acquired to win. We observe that LastSN-MAPPO shows much better and more stable final performance than MAPPO when evaluating under sparse reward configuration. However, the results on Corridor are a bit surprising. We talk about that in more detail in 
     Section \ref{section:res_smac}}
    \label{fig:smac_battles_won}
\end{figure*}

\section{Experimental Setup}
\label{section: experiment}
We use MAPPO as our on-policy multi-agent algorithm to perform all the evaluations. Implementation and configuration from \cite{epymarl} are used for all our experiments.
The actor consists of $3$ layers with GRU as the middle layer, and the critic uses $3$ layered feed-forward network. All the layers have $64$ neurons, and the hidden dimension of GRU is $64$. Adam \cite{kingmaAdamMethodStochastic2017} optimizer with a learning rate of $5 \times 10^{-4}$ is used for updating the network weights. Gradient clipping is applied to both the actor and critic gradients with a gradient norm $10$.
The weights of the actor and critic are shared across all agents\cite{yuMAPPOSurprisingEffectivenessPPO2022a}.

For learning the critic we use 10-step temporal difference learning rule. The actor is optimized using the standard PPO objective. We normalize the returns for critic for two of our variants, FullSN-MAPPO and LastSN-MAPPO.

We test three different variants with spectral normalization on critic and a standard MAPPO:
\begin{itemize}
    \item \emph{FullSN-MAPPO}: Spectral Normalization (SN) is applied on all critic layers.
    \item \emph{MidSN-MAPPO}: SN only applied on the second layer or the middle layer of the critic.
    \item \emph{LastSN-MAPPO}: SN applied to the final layer of the critic.
    \item \emph{MAPPO}: Standard MAPPO implementation with no spectral normalization.
\end{itemize}

\section{Results}
\label{section:results}
We empirically evaluate our results on two cooperative multi-agent benchmarks, multi-robot warehouse (RWARE) and starcraft multi-agent challenge (SMAC). We report our scores averaged across four seeds.

\subsection{RWARE Environment}

\textbf{RWARE} is a partially observable sparse reward benchmark introduced in \cite{epymarl}. It is a grid-world environment where the agents are rewarded for delivering the requested shelf from the warehouse. Agents can only observe a $3 \times 3$ grid surrounding themselves. We consider three different tasks which vary the grid size and the number of agents. This is a relatively simpler environment where a single agent can complete the task without any help from the other agents in the environment. This reflects in our results in Fig. \ref{fig:rware} where two variants, MAPPO and MidSN-MAPPO, show similar final performance, with MidSN-MAPPO being quicker to converge.

We compare three different RWARE environments with a varying number of agents and environment sizes. 
\begin{itemize}
    \item \emph{tiny-2ag} is the smallest map with two agents. We observe that the spectral normalized variant converges a bit faster comparatively.
    \item \emph{tiny-4ag} is the same as the previous map but with four agents. In this case, we do not see any significant difference between the two variants. Though our variant with normalized critic seems to converge a bit faster again.
    \item \emph{small-2ag} is a larger map with almost double the number of shelves in the environment with only two agents.
\end{itemize}
Overall in all three environments, we observe our variant to converge early, but the final performance is almost the same.

\begin{figure}
    \centering
    \includegraphics[width=0.36\textwidth]{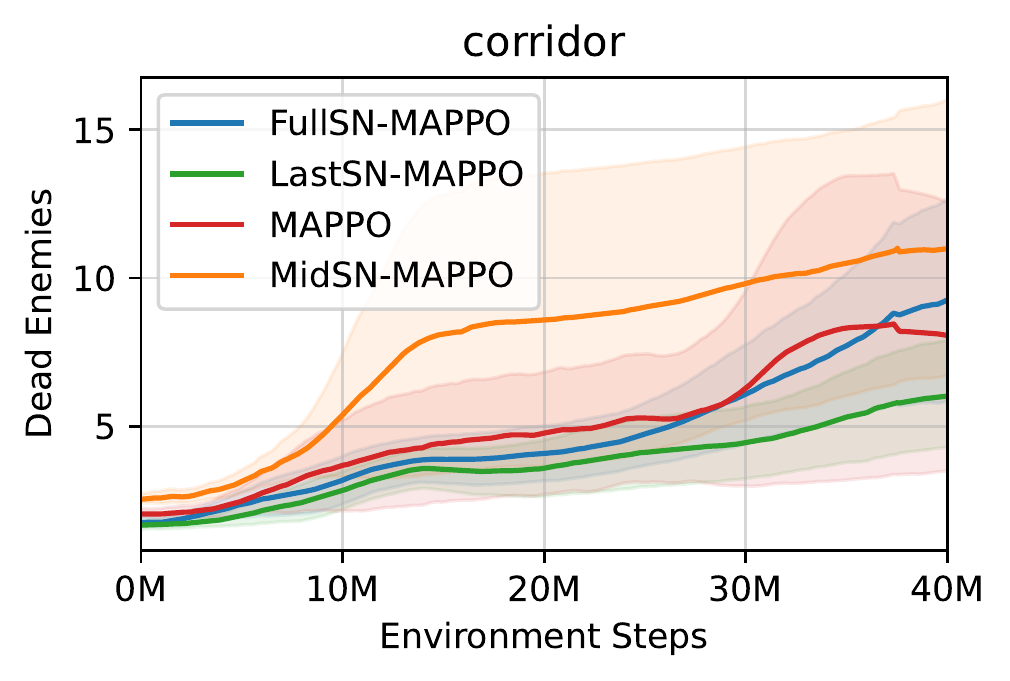}
    \caption{Comparing dead enemies through the training shows that MidSN-MAPPO is always ahead of MAPPO even though the final win rate is the same for the two variants, MidSN-MAPPO is able to kill more enemies even in the battles which are not a conclusive win.}
    \label{fig:corridor_deaths}
\end{figure}

\begin{figure}
    \centering
    \includegraphics[width=0.36\textwidth]{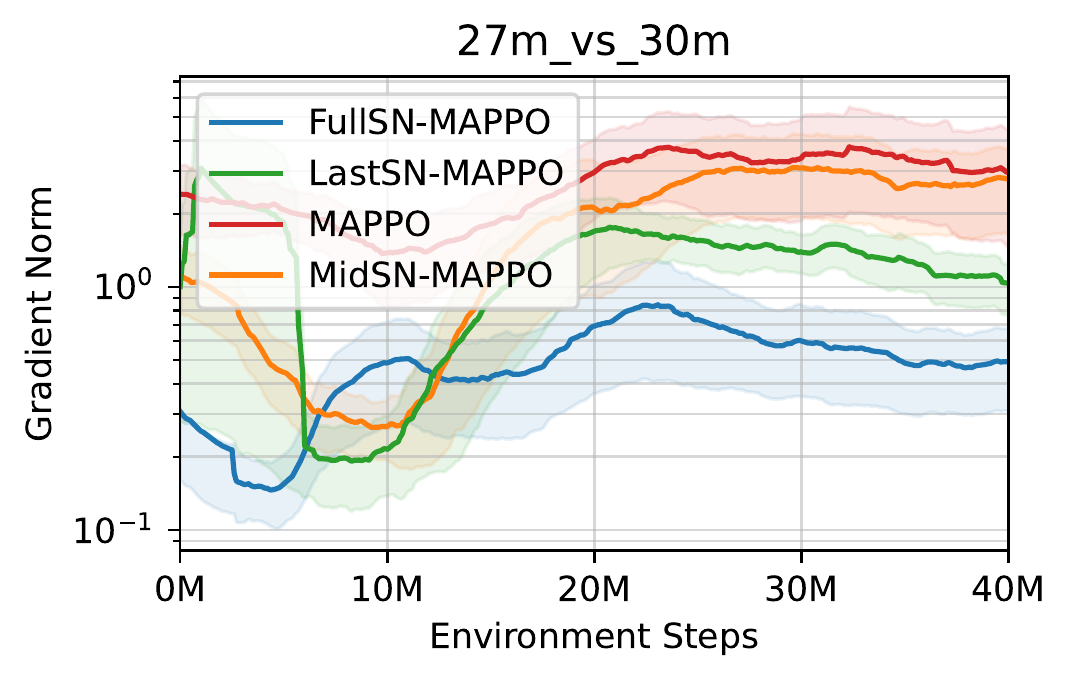}
    \includegraphics[width=0.36\textwidth]{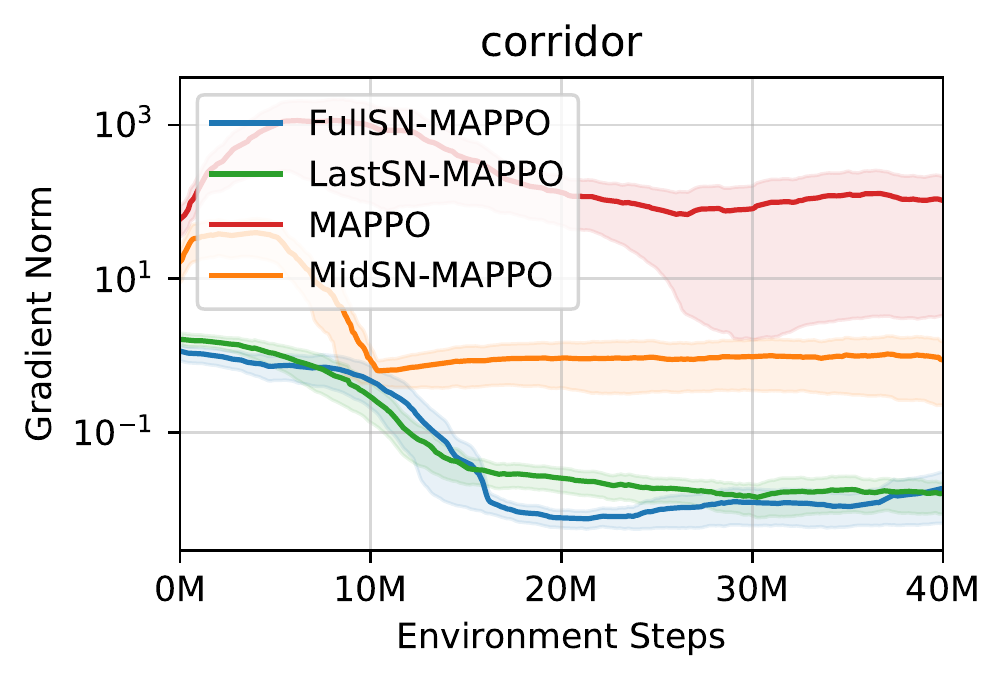}
    \caption{Gradient norm of the critic throughout the training. Even though MidSN-MAPPO shows improvement over MAPPO in both the environments, the reason for this performance gain is different in the two maps. In \emph{27m\_vs\_30m}, the critic gradients are stable for both the variants, still MidSN-MAPPO is better. In this case, the performance gain can be attributed to the optimization effects of SN on the critic. While in \emph{corridor}, SN helps stabilise the critic gradients, directly correlating to overall performance gain.}
    \label{fig:smac_critic}
\end{figure}

\subsection{SMAC Environment}
\label{section:res_smac}

\textbf{SMAC} is a benchmark based on the Starcraft II game. This environment consists of battle scenarios where a team of agents is controlled to defeat the enemy team, which uses fixed policies. This is also a partially observable environment where each agent only observes a fixed around itself for other agents. Here too, we consider three different tasks with a varying number of agents and unit types. The primary challenge in these tasks is learning optimal behaviour under partial observability and the large joint action space growing based on the number of agents. As we specifically wanted to evaluate the performance on sparse rewards, we propose a custom reward configuration where the agents are awarded rewards only in cases of death and win/loss. For each death in the ally team, a reward of $-10$ is awarded, and for each kill in the enemy team, a reward of $+10$ is awarded. Along with the death reward, a reward of $+200$ is awarded for winning the battle, killing all the enemy units, and similarly, a reward of $-200$ is awarded if all the units in the ally team die. We do not use rewards based on health loss due to attacks which are usually used.

We consider three \textbf{super-hard} scenarios from Starcraft Multi-Agent Challenge (SMAC) for our comparisons.
Each scenario evaluates different aspects of the environment.
\emph{3s5z\_vs\_3s6z} helps us evaluate the performance of imbalanced teams. We can observe that variants with spectral normalized critic gain significant performance compared to the standard critic variant. 
\emph{27m\_vs\_30m} has the largest ally team of 27 marines. In this scenario as well, we observe that our variant performs significantly better. This shows that our method can scale to a large number of agents. Even though spectral normalization constraints the critic, the shared weights can learn representation for many agents.
\emph{corridor} requires effective use of terrain features and block the choke point to avoid attacks from different directions. Subtle tactics like blocking the choke point to avoid attack from different directions as there is a considerable imbalance in the team since six friendly Zealots face 24 enemy Zerglings. All variants find it challenging to solve this environment consistently under sparse rewards. But still, the convergence of MidSN-MAPPO with normalized critic is quick compared to the standard variant. When we compare the number of dead enemies in Fig. \ref{fig:corridor_deaths}, we can see that MidSN-MAPPO is performing relatively better. Even though both the algorithms fail to have high win-rates due to slow regenerative ability of enemy Zerglings, which makes it difficult to kill them unless attacked continuously, we observe that MidSN-MAPPO is able to kill more enemies than MAPPO.

Fig. \ref{fig:smac_battles_won} compares the win rate on different SMAC scenarios under sparse rewards.
We can observe that all three SN variants perform better than the normal MAPPO on \emph{3s5z\_vs\_3s6z} and \emph{27m\_vs\_30m}. LastSN-MAPPO achieves the best final win-rate consistently across various seeds. This shows that regularizing the critic with spectral normalization does indeed help to learn under sparse reward scenarios. However, the results on \emph{corridor} paint a different picture. We observe that both the variants where SN is applied on the last layer of the critic underperform compared to the other two scenarios. 

Applying SN on the last layer of critic causes its output to be smooth \cite{sn-rl-opt}. However, the value function doesn't need to be smooth. That is, when the focal agent has more health and the enemy agent has relatively less health, the return will be highly positive, but just a slight difference in the health of the two agents leading to an enemy agent having higher health would lead to highly negative reward. The scenarios \emph{3s5z\_vs\_3s6z} and \emph{27m\_vs\_30m} where FullSN-MAPPO and LastSN-MAPPO perform well have open maps and there is a lot of place for the agents to move around. Hence the value function would be smooth. However, in \emph{corridor}, there are choke points that constraint the movements of the agents. This leads to non-smooth value function, which ultimately causes the failure of FullSN-MAPPO and LastSN-MAPPO on this scenario. It would be safe to conclude that applying SN on the final layer only helps when the value function is smooth. Otherwise, we have to restrict ourselves to not apply SN on the final layer of the critic.

To understand more about the effects of normalizing critic, we analyze the norm of the gradients of critic. Fig. \ref{fig:smac_critic} compares the gradient norm on two SMAC scenarios, \emph{27m\_vs\_30m} and \emph{corridor}. We observe that learning happens in both the maps, but there is a critic gradient explosion in the normal variant on \emph{corridor}. Notice that the plots are in $\log$ scale. This shows that regularising critic with spectral norm helps stabilize the learning in critic by stabilizing its gradients.

But another question that remains is what exactly causes the performance gain in \emph{27m\_vs\_30m}? As we observe, the gradient norm of both variants is almost in the same range. The performance gain, even when the gradient norm is not exploding, can be explained based on the effects of SN discussed in section \ref{section:optim_effect}. Let's look at the output and gradient equations of a three-layered fully-connected network. We observe that applying spectral normalization on a layer is equivalent to scaling the gradients of the complete network by the inverse of maximum spectral value $\rho^{-1}$. This scaling of the gradient effect acts as a step-size scheduler based on the spectral values of the regularised layers. Hence the performance gain in \emph{27m\_vs\_30m} can be attributed to the gradient scaling effect of SN. 

We can conclude from the above analysis that the benefits of applying spectral normalization to the critic are as follows
\begin{enumerate}
    \item Stabilise critic by constraining the gradients
    \item Optimization effect by scaling the gradient by the inverse of the maximum spectral value
    \item Better learning of smooth value functions by applying SN on the last critic layer
\end{enumerate}

However, it is important to note that even though SN can help in stabilizing the critic learning, it can only help up to an extent and under the conditions that the agent is able to reach some rewarding state by random exploration. In case of extremely sparse rewards, e.g., only win/loss reward in SMAC, it is extremely unlikely that the team of agents randomly stumbles upon a winning situation. As there is a very slim chance of getting an actual positive reward, there is no information presented to the critic that it can leverage. Hence stable critic helps only under the condition that the agent is able to reach rewarding states, but the reward signal might get suppressed by the noise from the untrained critic.

\section{Conclusion and Future Work}
We have investigated the challenges of sparse rewards in multi-agent environments and have empirically shown that regularising the critic with Spectral Normalization helps to learn a better policy. We show that in multi-agent sparse rewards scenarios, the benefits of applying SN are two folds, it restricts the irregularities in critic and stabilizes its gradients, and also changes the optimization dynamics by gradient scaling of the entire network. It is crucial to consider the smoothness of the value function of the environment when applying SN to the critic. Applying SN on the final layer of the critic when the value function is non-smooth hurts the performance. These observations highlight the importance of stable critic in MARL and show how SN can improve critic learning under challenging conditions.

As a future work, it would be interesting to explore the effects of spectral normalization when used on the actor network and using a deeper critic network. Moreover, our experiments were only limited to cooperative environments. Knowing how constraining the network with spectral normalization affects learning in general-sum games would give us more insight into its benefits in more general MARL settings. Additionally, we would also like to explore further the analytic relation between sample efficiency in RL \cite{mahajan2017symmetry} and the degree of spectral norm applied.

\section*{Acknowledgement}
This work was carried out under RIPPLE funding at the International Institute of Information Technology, Hyderabad, India. We thank the host institute for providing compute resources for an extended period of time, and for HPC resources funded under the RIPPLE grant. 

\bibliographystyle{IEEEtran}
\bibliography{IEEEexample}

\begin{thebibliography}{10}
\providecommand{\url}[1]{#1}
\csname url@samestyle\endcsname
\providecommand{\newblock}{\relax}
\providecommand{\bibinfo}[2]{#2}
\providecommand{\BIBentrySTDinterwordspacing}{\spaceskip=0pt\relax}
\providecommand{\BIBentryALTinterwordstretchfactor}{4}
\providecommand{\BIBentryALTinterwordspacing}{\spaceskip=\fontdimen2\font plus
\BIBentryALTinterwordstretchfactor\fontdimen3\font minus
  \fontdimen4\font\relax}
\providecommand{\BIBforeignlanguage}[2]{{%
\expandafter\ifx\csname l@#1\endcsname\relax
\typeout{** WARNING: IEEEtran.bst: No hyphenation pattern has been}%
\typeout{** loaded for the language `#1'. Using the pattern for}%
\typeout{** the default language instead.}%
\else
\language=\csname l@#1\endcsname
\fi
#2}}
\providecommand{\BIBdecl}{\relax}
\BIBdecl

\bibitem{mehta2023marljax}
\BIBentryALTinterwordspacing
K.~Mehta, A.~Mahajan, and P.~Kumar, ``marl-jax: Multi-agent reinforcement
  leaning framework for social generalization,'' 2023. [Online]. Available:
  \url{https://arxiv.org/abs/2303.13808}
\BIBentrySTDinterwordspacing

\bibitem{mahajan2022generalization}
A.~Mahajan, M.~Samvelyan, T.~Gupta, B.~Ellis, M.~Sun, T.~Rockt{\"a}schel, and
  S.~Whiteson, ``Generalization in cooperative multi-agent systems,''
  \emph{arXiv preprint arXiv:2202.00104}, 2022.

\bibitem{mahajan2019maven}
A.~Mahajan, T.~Rashid, M.~Samvelyan, and S.~Whiteson, ``Maven: Multi-agent
  variational exploration,'' in \emph{NeurIPS}, 2019, pp. 7611--7622.

\bibitem{gupta2020uneven}
T.~Gupta, A.~Mahajan, B.~Peng, W.~B{\"o}hmer, and S.~Whiteson, ``Uneven:
  Universal value exploration for multi-agent reinforcement learning,''
  \emph{arXiv:2010.02974}, 2020.

\bibitem{lowe2017multi}
R.~Lowe, Y.~Wu, A.~Tamar, J.~Harb, O.~P. Abbeel, and I.~Mordatch, ``Multi-agent
  actor-critic for mixed cooperative-competitive environments,'' in
  \emph{NeurIPS}, 2017, pp. 6379--6390.

\bibitem{qmix}
\BIBentryALTinterwordspacing
T.~Rashid, M.~Samvelyan, C.~S. de~Witt, G.~Farquhar, J.~Foerster, and
  S.~Whiteson, ``Qmix: Monotonic value function factorisation for deep
  multi-agent reinforcement learning,'' 2018. [Online]. Available:
  \url{https://arxiv.org/abs/1803.11485}
\BIBentrySTDinterwordspacing

\bibitem{son2019qtran}
K.~Son, D.~Kim, W.~J. Kang, D.~E. Hostallero, and Y.~Yi, ``Qtran: Learning to
  factorize with transformation for cooperative multi-agent reinforcement
  learning,'' \emph{arXiv:1905.05408}, 2019.

\bibitem{wang2020rode}
T.~Wang, T.~Gupta, A.~Mahajan, B.~Peng, S.~Whiteson, and C.~Zhang, ``Rode:
  Learning roles to decompose multi-agent tasks,'' \emph{arXiv:2010.01523},
  2020.

\bibitem{yu2021surprising}
C.~Yu, A.~Velu, E.~Vinitsky, Y.~Wang, A.~Bayen, and Y.~Wu, ``The surprising
  effectiveness of ppo in cooperative, multi-agent games,'' 2021.

\bibitem{pmlr-v139-mahajan21a}
A.~Mahajan, M.~Samvelyan, L.~Mao, V.~Makoviychuk, A.~Garg, J.~Kossaifi,
  S.~Whiteson, Y.~Zhu, and A.~Anandkumar, ``Tesseract: Tensorised actors for
  multi-agent reinforcement learning,'' in \emph{ICML}, vol. 139.\hskip 1em
  plus 0.5em minus 0.4em\relax PMLR, 2021, pp. 7301--7312.

\bibitem{yuMAPPOSurprisingEffectivenessPPO2022a}
\BIBentryALTinterwordspacing
C.~Yu, A.~Velu, E.~Vinitsky, J.~Gao, Y.~Wang, A.~Bayen, and Y.~Wu, ``The
  {{Surprising Effectiveness}} of {{PPO}} in {{Cooperative Multi-Agent
  Games}},'' 2022. [Online]. Available:
  \url{https://openreview.net/forum?id=YVXaxB6L2Pl}
\BIBentrySTDinterwordspacing

\bibitem{schulmanPPOProximalPolicyOptimization2017}
\BIBentryALTinterwordspacing
J.~Schulman, F.~Wolski, P.~Dhariwal, A.~Radford, and O.~Klimov, ``Proximal
  {{Policy Optimization Algorithms}},'' 2017. [Online]. Available:
  \url{http://arxiv.org/abs/1707.06347}
\BIBentrySTDinterwordspacing

\bibitem{sn-large-drl}
\BIBentryALTinterwordspacing
J.~Bjorck, C.~P. Gomes, and K.~Q. Weinberger, ``Towards {{Deeper Deep
  Reinforcement Learning}} with {{Spectral Normalization}},'' 2022. [Online].
  Available: \url{http://arxiv.org/abs/2106.01151}
\BIBentrySTDinterwordspacing

\bibitem{goodfellow2014}
I.~Goodfellow, J.~Pouget-Abadie, M.~Mirza, B.~Xu, D.~Warde-Farley, S.~Ozair,
  A.~Courville, and Y.~Bengio, ``Generative adversarial nets,'' in
  \emph{NeurIPS}, vol.~27, 2014.

\bibitem{florian2019}
F.~Schaefer and A.~Anandkumar, ``Competitive gradient descent,'' in
  \emph{NeurIPS}, H.~Wallach, H.~Larochelle, A.~Beygelzimer,
  F.~d\textquotesingle Alch\'{e}-Buc, E.~Fox, and R.~Garnett, Eds.,
  vol.~32.\hskip 1em plus 0.5em minus 0.4em\relax Curran Associates, Inc.,
  2019.

\bibitem{lars2017conopt}
L.~Mescheder, S.~Nowozin, and A.~Geiger, ``The numerics of gans,'' in
  \emph{NeurIPS}, ser. NIPS'17, Red Hook, NY, USA, 2017, p. 1823–1833.

\bibitem{kumar2023}
S.~K. Danisetty, S.~R. Mylaram, and P.~Kumar, ``Adaptive consensus optimization
  method for gans.''\hskip 1em plus 0.5em minus 0.4em\relax IJCNN, 2023.

\bibitem{sn-gan}
\BIBentryALTinterwordspacing
T.~Miyato, T.~Kataoka, M.~Koyama, and Y.~Yoshida, ``Spectral normalization for
  generative adversarial networks,'' in \emph{International Conference on
  Learning Representations}, 2018. [Online]. Available:
  \url{https://openreview.net/forum?id=B1QRgziT-}
\BIBentrySTDinterwordspacing

\bibitem{smac}
M.~Samvelyan, T.~Rashid, C.~S. {de Witt}, G.~Farquhar, N.~Nardelli, T.~G.
  Rudner, C.-M. Hung, P.~H. Torr, J.~Foerster, and S.~Whiteson, ``The
  {{StarCraft}} multi-agent challenge,'' vol.~4.\hskip 1em plus 0.5em minus
  0.4em\relax {International Foundation for Autonomous Agents and Multiagent
  Systems}, 2019.

\bibitem{epymarl}
\BIBentryALTinterwordspacing
G.~Papoudakis, F.~Christianos, L.~Sch{\"a}fer, and S.~V. Albrecht,
  ``Benchmarking {{Multi-Agent Deep Reinforcement Learning Algorithms}} in
  {{Cooperative Tasks}},'' 2021. [Online]. Available:
  \url{http://arxiv.org/abs/2006.07869}
\BIBentrySTDinterwordspacing

\bibitem{sunehagVDNValueDecompositionNetworksCooperative2017}
P.~Sunehag, G.~Lever, A.~Gruslys, W.~M. Czarnecki, V.~Zambaldi, M.~Jaderberg,
  M.~Lanctot, N.~Sonnerat, J.~Z. Leibo, K.~Tuyls, and T.~Graepel,
  ``Value-{{Decomposition Networks For Cooperative Multi-Agent Learning}},''
  \emph{arXiv:1706.05296}, 2017.

\bibitem{goukRegularisationNeuralNetworks2020}
\BIBentryALTinterwordspacing
H.~Gouk, E.~Frank, B.~Pfahringer, and M.~J. Cree, ``Regularisation of {{Neural
  Networks}} by {{Enforcing Lipschitz Continuity}},'' 2020. [Online].
  Available: \url{http://arxiv.org/abs/1804.04368}
\BIBentrySTDinterwordspacing

\bibitem{liuSimplePrincipledUncertainty2020}
\BIBentryALTinterwordspacing
J.~Z. Liu, Z.~Lin, S.~Padhy, D.~Tran, T.~{Bedrax-Weiss}, and
  B.~Lakshminarayanan, ``Simple and {{Principled Uncertainty Estimation}} with
  {{Deterministic Deep Learning}} via {{Distance Awareness}},'' 2020. [Online].
  Available: \url{http://arxiv.org/abs/2006.10108}
\BIBentrySTDinterwordspacing

\bibitem{yuMOPOModelbasedOffline2020}
\BIBentryALTinterwordspacing
T.~Yu, G.~Thomas, L.~Yu, S.~Ermon, J.~Zou, S.~Levine, C.~Finn, and T.~Ma,
  ``{{MOPO}}: {{Model-based Offline Policy Optimization}},'' 2020. [Online].
  Available: \url{http://arxiv.org/abs/2005.13239}
\BIBentrySTDinterwordspacing

\bibitem{sn-rl-opt}
F.~Gogianu, T.~Berariu, M.~Rosca, C.~Clopath, L.~Busoniu, and R.~Pascanu,
  ``Spectral {{Normalisation}} for {{Deep Reinforcement Learning}}: An
  {{Optimisation Perspective}},'' 2021.

\bibitem{andrychowiczHindsightExperienceReplay}
M.~Andrychowicz, D.~Crow, A.~Ray, J.~Schneider, R.~Fong, P.~Welinder,
  B.~McGrew, J.~Tobin, O.~P. Abbeel, and W.~Zaremba, ``Hindsight {{Experience
  Replay}},'' \emph{arXiv:2010.01523}.

\bibitem{miyatoSpectralNormalizationGenerative2022}
T.~Miyato, T.~Kataoka, M.~Koyama, and Y.~Yoshida, ``Spectral {{Normalization}}
  for {{Generative Adversarial Networks}},'' in \emph{ICLR}, 2022.

\bibitem{calrson2015}
D.~E. Carlson, E.~Collins, Y.-P. Hsieh, L.~Carin, and V.~Cevher,
  ``Preconditioned spectral descent for deep learning,'' in \emph{NeurIPS},
  vol.~28, 2015.

\bibitem{katyan2020}
S.~Katyan, S.~Das, and P.~Kumar, ``Two-grid preconditioned solver for bundle
  adjustment,'' in \emph{WACV}, 2020, pp. 3588--3595.

\bibitem{Das2021}
S.~Das, S.~Katyan, and P.~Kumar, ``A deflation based fast and robust
  preconditioner for bundle adjustment,'' in \emph{WACV}, January 2021, pp.
  1782--1789.

\bibitem{das2020}
------, ``Domain decomposition based preconditioned solver for bundle
  adjustment,'' in \emph{Computer Vision, Pattern Recognition, Image
  Processing, and Graphics}, 2020.

\bibitem{kumar2013multi}
P.~Kumar, K.~Meerbergen, and D.~Roose, ``Multi-threaded nested filtering
  factorization preconditioner,'' in \emph{Applied Parallel and Scientific
  Computing}, vol. 7782, Springer.\hskip 1em plus 0.5em minus 0.4em\relax
  Springer, Berlin, Heidelberg, 2013, pp. 220--234.

\bibitem{kumar2014aggregation}
P.~Kumar, ``Aggregation based on graph matching and inexact coarse grid solve
  for algebraic two grid,'' \emph{International Journal of Computer
  Mathematics}, vol.~91, no.~5, pp. 1061--1081, 2014.

\bibitem{kumar2016relaxed}
P.~Kumar, L.~Grigori, F.~Nataf, and Q.~Niu, ``On relaxed nested factorization
  and combination preconditioning,'' \emph{International Journal of Computer
  Mathematics}, vol.~93, no.~1, pp. 179--199, 2016.

\bibitem{li2015preconditioned}
C.~Li, C.~Chen, D.~Carlson, and L.~Carin, ``Preconditioned stochastic gradient
  langevin dynamics for deep neural networks,'' 2015.

\bibitem{qiao2019}
Y.~Qiao, B.~P.~F. Lelieveldt, and M.~Staring, ``An efficient preconditioner for
  stochastic gradient descent optimization of image registration,'' \emph{IEEE
  Transactions on Medical Imaging}, vol.~38, no.~10, pp. 2314--2325, 2019.

\bibitem{kingmaAdamMethodStochastic2017}
\BIBentryALTinterwordspacing
D.~P. Kingma and J.~Ba, ``Adam: {{A Method}} for {{Stochastic Optimization}},''
  2017. [Online]. Available: \url{http://arxiv.org/abs/1412.6980}
\BIBentrySTDinterwordspacing

\bibitem{mahajan2017symmetry}
A.~Mahajan and T.~Tulabandhula, ``Symmetry learning for function approximation
  in reinforcement learning,'' \emph{arXiv preprint arXiv:1706.02999}, 2017.

\end{thebibliography}

\end{document}